\pdfoutput=1

\documentclass[11pt]{article}

\usepackage[final]{acl}

\usepackage{times}
\usepackage{latexsym}
\usepackage{float}
\usepackage[T1]{fontenc}

\usepackage[utf8]{inputenc}

\usepackage{microtype}

\usepackage{inconsolata}

\usepackage{graphicx}

\usepackage{booktabs}
\usepackage{multirow}
\usepackage{hyperref}
\usepackage{tcolorbox}
\usepackage{amsmath}

\newcommand\blfootnote[1]{%
  \begingroup
  \renewcommand\thefootnote{}\footnote{#1}%
  \addtocounter{footnote}{-1}%
  \endgroup
}

\title{\textsc{ETHIC}: Evaluating Large Language Models on Long-Context Tasks with High Information Coverage}

\author{Taewhoo Lee$^{1,3}$ \quad Chanwoong Yoon$^1$ \quad Kyochul Jang$^2$  \quad Donghyeon Lee$^{1,3}$ \quad Minju Song$^1$  \\
\textbf{Hyunjae Kim$^{1,\dagger}$} \quad \textbf{Jaewoo Kang$^{1,3,\dagger}$}\\
$^1$Korea University \quad $^2$Seoul National University \quad $^3$AIGEN Sciences \\
\texttt{\{taewhoo,cwyoon99,dong9733,thdalswn99\}@korea.ac.kr}\\
\texttt{kyochul@snu.ac.kr} \quad \texttt{\{hyunjae-kim,kangj\}@korea.ac.kr}
}

\begin{document}
\maketitle

\newcommand{\ours}{\textsc{ETHIC}}

\newcommand{\draftonly}[1]{#1}

\newcommand{\draftcomment}[3]{\draftonly{\textcolor{#2}{{\textbf{[#3 --\textsc{#1}]}}}}}

\newcommand{\todo}[1]{\draftcomment{TODO}{red}{#1}}
\newcommand{\mujeen}[1]{\draftcomment{mujeen}{cyan}{#1}}
\newcommand{\chanhwi}[1]{\draftcomment{chanhwi}{red}{#1}}
\newcommand{\hyunjae}[1]{\draftcomment{hyunjae}{teal}{#1}}
\newcommand{\sihyeon}[1]{\draftcomment{sihyeon}{brown}{#1}}
\newcommand{\jiwoo}[1]{\draftcomment{jiwoo}{purple}{#1}}

\begin{abstract}
Recent advancements in large language models (LLM) capable of processing extremely long texts highlight the need for a dedicated evaluation benchmark to assess their long-context capabilities.
However, existing methods, like the needle-in-a-haystack test, do not effectively assess whether these models fully utilize contextual information, raising concerns about the reliability of current evaluation techniques.
To thoroughly examine the effectiveness of existing benchmarks, we introduce a new metric called information coverage (IC), which quantifies the proportion of the input context necessary for answering queries. Our findings indicate that current benchmarks exhibit low IC; although the input context may be extensive, the actual usable context is often limited.
To address this, we present \ours, a novel benchmark designed to assess LLMs' ability to leverage the entire context. 
Our benchmark comprises 1,986 test instances spanning four long-context tasks with high IC scores in the domains of books, debates, medicine, and law.
Our evaluations reveal significant performance drops in contemporary LLMs, highlighting a critical challenge in managing long contexts.
Our benchmark is available at \href{https://github.com/dmis-lab/ETHIC}{https://github.com/dmis-lab/ETHIC}.
\end{abstract}
\blfootnote{\textsuperscript{$\dagger$}Corresponding authors.}

\section{Introduction}
\label{section:introduction}

The field of natural language processing (NLP) has made remarkable progress in developing models that can manage much longer texts. 
While earlier Transformer-based models~\cite{vaswani2017attention} could only process 512 tokens at a time~\cite{kenton2019bert,raffel2020exploring}, modern large language models (LLM) have achieved a significant breakthrough, now capable of handling documents with up to two million tokens~\cite{reid2024gemini}. 
In light of these advancements, recent efforts have focused on establishing benchmarks and tasks specifically designed to evaluate the performance of these long-context models~\cite{shaham-etal-2023-zeroscrolls,hsieh2024ruler}.

\begin{figure}[t!]
\centering
\includegraphics[width=\columnwidth]{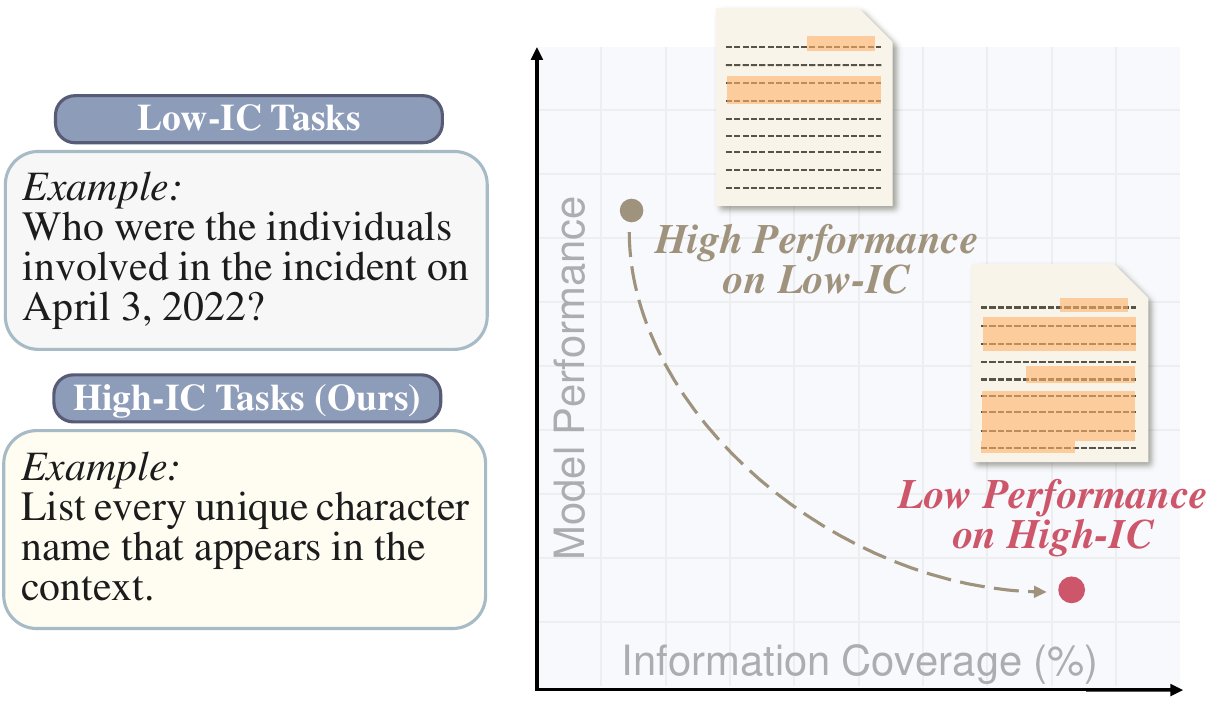}
\caption{
The variation in model performance with the level of information coverage (IC). 
Unlike low-IC tasks, which focus on specific parts of the input context, our benchmark features new high-IC tasks that demand the full utilization of all available information, posing a significant challenge for long-context models.
}
\label{fig:motivation}
\end{figure}

However, current long-context benchmarks often face challenges in assessing whether models are fully utilizing the information available in the provided context. 
One common research method, known as the needle-in-a-haystack test~\cite{kamradt2023needle}, aims to identify a specific piece of information within a lengthy context. 
However, excelling in these tasks does not guarantee that the model has effectively processed all the available information. 
Since the relevant information typically constitutes only a small portion of the entire text, much of the surrounding content is often irrelevant to the query.
While several datasets have proposed tasks involving multiple key pieces of information scattered throughout the provided context~\cite{dong-etal-2024-bamboo, li-etal-2024-loogle, wang2024leave}, they still do not fully encompass the entire context. 
This raises concerns about whether models are adequately evaluated on their ability to fully incorporate the entire context length~\cite{goldman2024really}.

To this end, we propose~\ours, a suite of long-context tasks specifically designed to assess whether LLMs can fully utilize the provided information. 
Our benchmark encompasses four distinct domains: books, debates, medicine, and law, each containing a set of tasks that require the use of all relevant information in the context to arrive at a solution.
To measure this capability, we introduce the concept of \textit{information coverage} (IC), which quantifies the proportion of the input context required to answer a query.
Figure~\ref{fig:motivation} shows that model performance significantly declines for high-IC tasks in our benchmark compared to low-IC tasks, even when the same input contexts are used.

We evaluated the latest LLMs that support at least 128k tokens, along with several training-free methods using our benchmark.
Our findings revealed that recent models perform poorly across all tasks and domains, even when utilizing recent frameworks proposed for efficient long-context processing~\cite{xiao2024efficient,xiao2024infllm,qian2024memorag}. 
This highlights a significant challenge for high-IC tasks and the need for further research in this area.
We also conducted detailed analyses, comparing the performance gap between our task and traditional low-IC tasks, as well as identifying which parts of the input context the models typically fail to address.
In summary, our contributions include:
\begin{itemize}
    \item We introduce a new metric called information coverage (IC) to measure the proportion of input context required to answer a query.
    \item We propose~\ours, the first benchmark of its kind, designed to assess whether LLMs can fully process the provided information.
    Our benchmark requires a higher IC than existing benchmarks, which presents a new challenge for the latest LLMs.
    \item We conduct a comprehensive analysis of how LLMs perform in high-IC tasks, establishing a foundation for future research on the development of advanced long-context models.  
\end{itemize}

\section{Preliminaries}
\label{section:preliminaries}

In this section, we provide an overview of recent long-context LLMs and the benchmarks currently used to evaluate them (see Sections~\ref{subsec:llms}~and~\ref{subsec:benchmarks}). 
We also present a formal description of \textit{information coverage} and explain how our benchmark differs from existing ones, emphasizing the new aspects of LLMs that we aim to evaluate (see Section~\ref{subsec:information_coverage}).

\subsection{Long-Context LLMs}
\label{subsec:llms}


Pre-training LLMs on long contexts requires significant computational resources. Early open-source models, such as LLaMA~\cite{touvron2023llama}, could manage input lengths of about 2K tokens, while LLaMA 2~\cite{touvron2023llama2} increased this limit to 4K tokens. Even commercial models like GPT-3.5 initially supported input lengths of 16K tokens. 
More recent models have vastly improved these capabilities, now supporting input lengths that range from 128K~\cite{dubey2024llama} to two million tokens~\cite{reid2024gemini}.
However, the techniques employed to achieve this efficiency, along with the actual amount of input that models can effectively utilize, remain largely disclosed.

Several fine-tuning techniques have been explored to effectively extend the context window of pre-trained LLMs~\cite{zhu2023pose, chen2024longlora, peng2024yarn}. 
For instance, \citet{chen2023extending}~noted that simply training a model on longer contexts is both computationally intensive and often ineffective. They proposed a position interpolation method as a more efficient solution. 
Furthermore, approaches to extend the context window during inference—without requiring additional training—have also been investigated~\cite{pmlr-v235-jin24b, xiao2024infllm, xiao2024efficient, han2024lm}.

\begin{table*}[t]
\centering
\footnotesize
\begin{tabular}{lcccc}
\toprule
\textbf{Benchmark} & \begin{tabular}[c]{@{}c@{}}\textbf{Newly}\\ \textbf{Curated}\end{tabular} & \begin{tabular}[c]{@{}c@{}}\textbf{Input Text}\\ \textbf{Structure}\end{tabular} & \begin{tabular}[c]{@{}c@{}}\textbf{Document}\\ \textbf{Relevance}\end{tabular} & \begin{tabular}[c]{@{}l@{}}\textbf{Information}\\ \textbf{Coverage (\%)}\end{tabular} \\
\midrule
NIAH~\cite{kamradt2023needle} & Yes & Multi & Low &  N/A\\
RULER~\cite{hsieh2024ruler} & Partial & Multi & Low & N/A \\
Counting-Stars~\cite{song2024countingstars} & Yes & Multi & Low & N/A \\
\midrule
ZeroSCROLLS~\citep{shaham-etal-2023-zeroscrolls} & No & Single \& Multi & High & 56.1  \\
L-Eval~\citep{an-etal-2024-l} & Partial & Single \& Multi & High & 35.4 \\
InfiniteBench~\citep{zhang-etal-2024-bench} & Yes & Multi & Mixed & 16.5 \\
BAMBOO~\citep{dong-etal-2024-bamboo} & Yes & Single & High & 41.4\\
Loong~\citep{wang2024leave} & Yes & Multi & High & 14.4 \\
LooGLE~\citep{li-etal-2024-loogle} & Yes & Single & High & 9.6 \\
\midrule
\ours~(\textbf{Ours}) & Yes & Single \& Multi & High & 91.0 \\
\bottomrule
\end{tabular}
\caption{
Comparison of existing long-context benchmarks and our dataset. 
``Newly Curated'' indicates whether the input text and queries/instructions are reused from existing datasets or newly created. ``Input Text Structure'' specifies whether the input context consists of a single document or multiple documents. 
``Document Relevance'' assesses whether the different documents in multi-document tasks are unrelated and noisy, or if they are connected and coherent.
``Information Coverage'' quantifies the amount of information within the input context that is necessary to answer the query.
Note that information coverage is marked as "N/A" if the proportion of required information varies depending on custom settings.  
Please refer to Sections~\ref{section:preliminaries} and~\ref{section:benchmark} for information on the datasets.
}
\label{table:benchmark_comparison}
\end{table*}

\subsection{Long-Context Benchmarks}
\label{subsec:benchmarks}
Researchers have been evaluating how well long-context LLMs handle extensive text. 
A common approach is the needle-in-a-haystack (NIAH) task, where the goal is to locate key information (the ``needle'') within a large volume of text (the ``haystack'')~\cite{kamradt2023needle,NEURIPS2023_ab05dc8b}. 
Some studies manipulate the number of needles and the haystack's length to increase the complexity~\cite{hsieh2024ruler,song2024countingstars}. 
Additionally, several studies have adapted traditional NLP tasks--such as retrieval, single-document QA, and summarization--to serve as long-context evaluation scenarios~\cite{shaham-etal-2023-zeroscrolls, an-etal-2024-l, bai-etal-2024-longbench,zhang-etal-2024-bench}. 
Some benchmarks target long-dependency or multi-hop reasoning and distribute information throughout the context~\cite{dong-etal-2024-bamboo, li-etal-2024-loogle, wang2024leave}.
However, these benchmarks often utilize contexts in which a significant portion of the text is irrelevant to the query, and only a small segment contains useful information. 
Consequently, they do not fully assess how well LLMs understand and integrate different parts of the given context.

In contrast, our benchmark requires models to make extensive use of the provided context. 
To measure the necessary amount of information, we introduce a metric called information coverage, which is explained in detail in Section~\ref{subsec:information_coverage}.
Table~\ref{table:benchmark_comparison} illustrates the differences between existing benchmarks and our proposed benchmark.

\begin{figure*}[t]
\includegraphics[width=\textwidth]{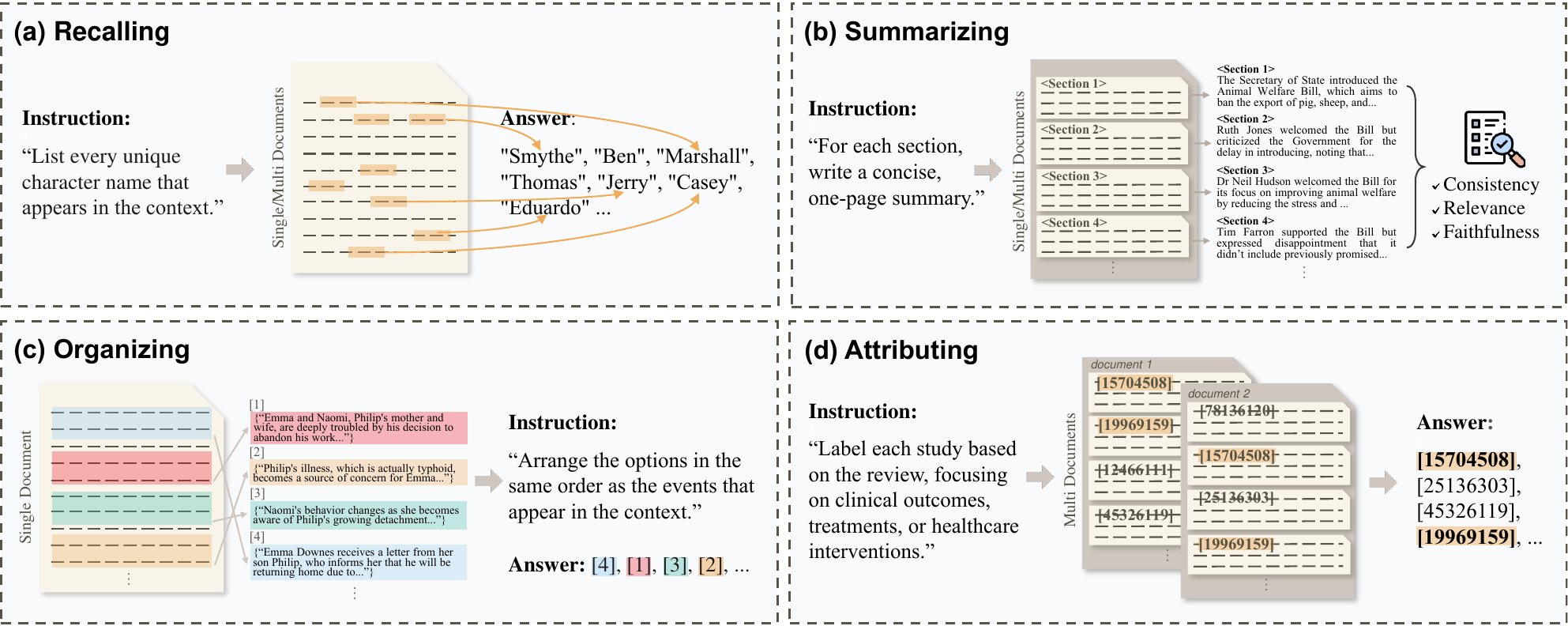}
\caption{
Overall description of~\ours.
Our benchmark includes four tasks: (a) the recalling task involves identifying specific types of entities in the text, (b) the summarizing task involves writing a summary for each section of the input, (c) the organizing task involves arranging mixed contents in the correct order, and (d) the attributing task focuses on identifying the underlying point of view within medical studies or legal documents.
}
\label{figure:subtasks}
\end{figure*}

\subsection{Information Coverage}
\label{subsec:information_coverage}

Let $\mathcal{D}$ = \{($\mathbf{C}_i$,$\mathbf{q}_i$\,$\mathbf{a}_i$)\}$_{i=1}^N$ be a dataset, where $\mathbf{C}_i$, $\mathbf{q}_i$, and $\mathbf{a}_i$ represent the $i$-th input long context, query, and output (answer), respectively.
The input context can be a single document, such as a book, or a collection of related documents, like a set of papers on a similar research topic.
We omit the subscript $i$ for simplicity.
We divide the input context into chunks, each with a length of up to 512, denoted as $\{\mathbf{c}_1, \dots, \mathbf{c}_{T}\}$. 
For each data example, we calculate the IC score by using an evaluator model to determine whether each text chunk is (potentially) necessary for answering the query, as follows:
\begin{equation}
    s(\mathbf{C}, \mathbf{q}) = \frac{1}{T}\sum^{T}_{j=1} \mathcal{M}(\mathbf{c}_j,\mathbf{q}),
\end{equation}
where $s$ is the IC scoring function, $\textbf{q}$ is the query, and $\mathcal{M}$ represents the evaluator model that returns 1 if the text chunk should be taken into account and 0 otherwise.
The IC score for the entire dataset is calculated as the average of all individual IC scores from the examples.
We used GPT-4o as the evaluator because of its higher consistency compared to other models.
Please refer to Appendix~\ref{sec:appendixA} for the detailed prompt.

A few concurrent studies have also sought to establish the criteria for evaluating long-context LLMs. 
\citet{goldman2024really}~defined the aspect of ``scope'' as ``how much necessary information is there to find?'', which is similar to our definition.
However, they did not provide a specific metric and systematically evaluate existing benchmarks. 
In contrast, we present a novel approach for measuring the quantity of information, marking the first time this has been done in this area.

\section{\ours}
\label{section:benchmark}
In this section, we outline the document collection process (see Section~\ref{subsec:corpus_selection}) and the task construction process (see Section~\ref{subsec:task_construction}) in detail. Additionally, we describe the label annotation process for each task in Section~\ref{subsec:annotation}. 

\subsection{Corpus Selection}
~\label{subsec:corpus_selection}
We selected four popular domains and gathered publicly available documents online without license restrictions for research purposes: books, debates, medicine, and law. 
The first two domains are used for single-document settings, while the latter two are for multi-document settings.


\paragraph{Books} 
Book-sourced corpora have been widely used across various benchmarks~\cite{kocisky-etal-2018-narrativeqa, kryscinski-etal-2022-booksum, chang2024booookscore}. 
We collected 100 English books from Project Gutenberg,\footnote{\url{https://www.gutenberg.org}} all of which are no longer under copyright as of 2024. 

\paragraph{Debates} 
We collected 229 debate transcripts from Hansard online,\footnote{\url{https://hansard.parliament.uk/}} which provides up-to-date records of debates held in the UK Parliament.
Parliamentary debates cover a wide range of subjects, including political issues and legislative proposals. 
We manually selected debates by filtering out those that are either short (fewer than 10k tokens) or not considered debates, such as maiden speeches.

\paragraph{Medicine} 
We collected 230 samples from the test 
set of MS\^{}2~\cite{deyoung-etal-2021-ms}, a multi-document summarization dataset in which each sample comprises relevant medical abstracts used for systematic reviews. 
We excluded any samples with fewer than 10K tokens.

\paragraph{Law} 
We gathered legal cases using the API from CourtListener\footnote{\url{https://www.courtlistener.com}}, which provides up-to-date legal documents for research purposes. 
We grouped each target case with up to 15 related cases that cite it, resulting in a total of 103 samples.

\begin{table*}[t]
\centering
\footnotesize
\begin{tabular}{lllccc}
\toprule
\textbf{Task} & \begin{tabular}[c]{@{}c@{}}\textbf{Cognitive}\\ \textbf{Process}\end{tabular} & \textbf{Domain} & \multicolumn{1}{c}{\textbf{\# Instances}} & \multicolumn{1}{c}{\textbf{\# Avg. Tokens}} & \begin{tabular}[c]{@{}l@{}}\textbf{Information}\\ \textbf{Coverage (\%)}\end{tabular} \\
\midrule 
\multirow{1}{*}{Recalling} & Remember & Books & 100 & 76,812 & 93.8 \\
 && Debates & 229 & 28,524 & 82.1 \\
 && Medicine ($\dagger$) & 230 & 29,037 & 86.6 \\
 && Law ($\dagger$) & 103 & 59,938 & 85.4 \\
\midrule 
Summarizing & \multirow{1}{*}{Understand} & Books & 100 & 76,893 & 92.3 \\
 && Debates & 229 & 28,598 & 97.7 \\
 && Medicine ($\dagger$) & 230 & 29,065 & 91.7\ \\
 && Law ($\dagger$) & 103 & 59,970 & 98.1 \\
\midrule
Organizing & \multirow{1}{*}{Analyze} & Books & 100 & 94,403 & 74.4 \\
 && Debates & 229 & 35,751 & 87.7 \\
\midrule
Attributing & Analyze & Medicine ($\dagger$) & 230 & 29,166 & 97.5 \\
 && Law ($\dagger$) & 103 & 56,905  & 82.5 \\
\bottomrule
\end{tabular}
\caption{
A summary of our dataset construction.
~\ours~covers four domains—books, debates, medicine, and law—with a total of 1,986 test instances.
In the medicine and law domains (marked with $\dagger$), the inputs consist of multiple documents, while the inputs in the books and debates domains consist of a single long document. We randomly sampled 50 instances per domain and task to report information coverage. 
}
\label{table:data_statistics}
\end{table*}

\subsection{Task Construction}
\label{subsec:task_construction}

When designing tasks, we prioritized two key aspects: (1) maximizing the use of the provided context and (2) ensuring they are grounded in well-defined categorization standards.
By emphasizing these criteria, we enhanced the clarity and effectiveness of our benchmark, distinguishing it from existing long-context benchmarks.
To achieve this, we drew on insights from \citet{Anderson2000ATF}, an authoritative source in the education field that provides a systematic approach to classifying educational objectives. 
From this framework, we adopted the following three distinct cognitive categories that align with our goals.
(i) Remember: this cognitive process involves retrieving relevant information from the provided context in its original form. 
It serves as a crucial step for addressing more complex tasks that require integrating the knowledge gained during this process.
(ii) Understand: this category involves constructing meaning by interpreting and making sense of the knowledge obtained from the provided context.
(iii) Analyze: this process entails breaking down the provided context into its constituent parts and examining their relationships.
Based on these categories, we developed four tasks--recalling, summarizing, organizing, and attributing--each corresponding to one of the three categories.
Figure~\ref{figure:subtasks} illustrates the four tasks included in~\ours.

\paragraph{Recalling}
In this task, models should retrieve all specific types of entity mentions from the input context, similar to the named entity recognition task~\cite{sang2003introduction}.
This includes identifying character names from books, names of individuals from debates, numbers of patients or populations from medical studies, and legal references from legal cases, all of which consistently appear throughout the document. To eliminate ambiguity in the answer format, models are guided to return a list of single words or numbers only.

\paragraph{Summarizing} 
This task requires the model to summarize and rephrase the key ideas from the given context.
While current LLMs are known to excel at traditional summarization tasks~\cite{pu2023summarization}, evaluating this ability under long-context settings remains a challenge~\cite{wu-etal-2024-less}. 
We extended the existing summarization task to a high-IC setting. 
Instead of summarizing the long context all at once, we divided it into smaller text chunks and required models to summarize each chunk individually (e.g., sections in books).
This encourages models to capture essential information from each section without missing important details.

\paragraph{Organizing} 
In this task, models are provided with the entire context along with summaries of each chunk in a random order. Models  should then rearrange these summaries into their original sequence. The output consists of the document IDs in the correct order.
Conventional tasks generally require filling in missing parts in the correct order~\cite{wang-etal-2024-ada} or reordering a limited number of information chunks~\cite{dong-etal-2024-bamboo}, all of which can be solved by attending to specific areas instead of the entire context. 
Our task prevents these potential bypasses by instructing models to organize summary chunks that altogether represent the entire context. 
This task applies only to the single-document setting.

\paragraph{Attributing} 
Unlike the organizing task, this task applies only to the multi-document setting.
Models are tasked with inferring the underlying point of view within the given context. 
In the medicine domain, models receive a set of abstracts used in the same systematic review, along with a ``background'' paragraph.
The background section is taken from a different target review, and some of these abstracts are also referenced in that review. 
The models must then identify the IDs of the medical abstracts that were included in the target review.
For the law domain, we first grouped multiple pages from each legal case into segments. 
Models are given a target case, along with a set of segments from other cases that cite the target case. 
Note that the specific citations are masked using a citation mark. Models must go over each segment and check whether the context surrounding each citation mark aligns with the target case.
Overall, this task involves actively integrating information from each document and understanding the reasoning behind citing a particular study or case.

\subsection{Annotation Process}
This section provides details of our annotation process for each task, except for the summarizing task, where we adopted a reference-free method to evaluate the generated summaries~\cite{liu-etal-2023-g}. 
Table~\ref{table:data_statistics} provides a summary of our benchmark.   

\label{subsec:annotation}

\paragraph{Recalling} 

Since annotating long contexts all at once can reduce accuracy, we processed each context by dividing it into smaller chunks (up to 1,024 tokens). 
We used GPT-4o to annotate each small chunk, and merged the labels without duplicates to obtain the final label set for the full input document. We initially instructed GPT-4o to review its own answers, but empirically found that this process often led the model to misjudge correct answers as incorrect. We manually reviewed the accuracy of the model’s annotations by examining 100 chunk-label pairs from each domain and found that they were highly accurate (see Appendix~\ref{appendix:appendixC} for details).




\paragraph{Organizing} 
Generating labels for this task involves generating multiple summary chunks from the original context and shuffling them into random orders. 
Using the same small chunks used when annotating the recalling task, we prompted GPT-4o to briefly summarize each chunk with up to five sentences, and then we randomly shuffled them.

\paragraph{Attributing}

For the medical domain, each sample (i.e., a set of medical abstracts used in the same systematic review) was inspected to identify the PMIDs of abstracts that were also included in another sample, which became the label set. If multiple samples contained overlapping abstracts, we chose the one with the highest overlap. 
If no abstract was used in any other sample, we randomly selected background paragraphs, and the label set was labeled as ``none.''
For the law domain, each sample included one target case and multiple citing cases. 
We first generated a summary of the target case using GPT-4o. 
Then, for each page of a citing case, the model was instructed to identify all spans referring to the target case based on its title and summary. 
Any spans referring to the target case were replaced with target citation markers (``[TARGET CITATION]''), while references to other cases were replaced with generic citation markers (``[CITATION]''). 
Finally, we grouped pages into segments, identified the segment IDs containing target citation markers, and replaced every target citation marker with a generic citation marker.

\section{Experiments}
\label{section:experiments}

\begin{table*}[t]
\centering
\footnotesize
\begin{tabular}{lcccc}
\toprule
\multirow{3}{*}{\textbf{Model}} & \multicolumn{1}{c}{\textbf{Recalling}} & \multicolumn{1}{c}{\textbf{Summarizing}} & \multicolumn{1}{c}{\textbf{Organizing}} & \multicolumn{1}{c}{\textbf{Attributing}} \\
\cmidrule(lr){2-2} \cmidrule(lr){3-3} \cmidrule(lr){4-4} \cmidrule(lr){5-5}
 & \multicolumn{1}{c}{\textbf{F1 (\%)}} & \multicolumn{1}{c}{\textbf{Score (1-5)}} & \multicolumn{1}{c}{\textbf{LCS (\%)}} & \multicolumn{1}{c}{\textbf{F1 (\%)}} \\
\midrule
\multicolumn{5}{l}{\textit{Proprietary}} \\
\midrule
Gemini Pro 1.5~\cite{reid2024gemini} & \textbf{69.1} & 2.9 & \textbf{54.5} & 39.4 \\
GPT-4o~\cite{openai2024gpt4o} & 49.5 & \textbf{3.1} & 39.0 & \textbf{41.3} \\
GPT-4o mini~\cite{openai2024gpt4o} & 32.3 & 2.7 & 21.7 & 30.5 \\
\midrule
\multicolumn{5}{l}{\textit{Open-Source}} \\
\midrule
Qwen2.5-72B-Instruct~\cite{yang2024qwen2} & \textbf{45.4} & \textbf{3.0} & \textbf{27.9} & \textbf{45.1} \\
Llama-3.1-70B-Instruct~\cite{dubey2024llama} & 37.7 & 2.4 & 25.7 & 41.1 \\
\midrule
Qwen2.5-7B-Instruct~\cite{yang2024qwen2} & 14.3 & 2.7 & 19.0 & 24.2 \\
Llama-3.1-8B-Instruct~\cite{dubey2024llama} & 18.0 & 2.3 & 20.2 & 28.8 \\
GLM4-9B-Chat~\cite{glm2024chatglm} & 18.3 & 2.3 & 22.1 & 28.2 \\
Phi-3.5-mini-instruct (3.8B)~\cite{abdin2024phi} & 11.7 & 2.2 & 15.9 & 25.9 \\
\midrule
\multicolumn{5}{l}{\textit{Training-Free Methods (built upon Llama3.1-8B-Instruct)}} \\
\midrule
StreamingLLM~\cite{xiao2024efficient} & 15.8 & 1.6 & 1.7 & 10.6 \\
InfLLM~\cite{xiao2024infllm} & 17.0 & 1.8 & 13.8 & 12.7 \\
MemoRAG~\cite{qian2024memorag} & 16.8 & 1.7 & 22.1 & 16.7 \\
\bottomrule
\end{tabular}
\caption{
The performance of models and training-free methods on~\ours. 
The best scores are highlighted in bold.
We used instruction-tuned versions of open-source models.
Please refer to Section~\ref{subsec:metrics} for the details of the metrics.
}
\label{table:main_results}
\end{table*}
In this section, we outline different metrics used to evaluate each task (see Section~\ref{subsec:metrics}). We introduce baseline models and methods in Section~\ref{subsec:baselines}, and provide the results in Section~\ref{subsec:results}.

\subsection{Metrics}
\label{subsec:metrics}

For the recalling task, we used the F1 score to evaluate how well the predicted entities matched the ground truth.
For the summarizing task, we followed the approach of~\citet{wu-etal-2024-less}, prompting GPT-4o to rate the generated summaries on consistency, relevance, and faithfulness, using a scale from 1 to 5. 
The final score was calculated by multiplying the probability by the assigned score, similar to the method used by~\citet{liu2023g}.
For the organizing task, we found that models can hardly obtain any scores when evaluated using Exact Match. Therefore, we used Longest Common Subsequence (LCS), which measures the proportion of the longest matching subsequence between the prediction and the ground truth, relative to the total sequence length. The subsequences do not need to be consecutive.
Lastly, for the attributing task, we used the F1 score to compare the predicted document IDs with the ground truth.

\subsection{Baselines}
\label{subsec:baselines}
\paragraph{Long-Context Models}
We used the current best LLMs that support a context window of over 128k tokens on~\ours. 
This included three powerful proprietary models--Gemini Pro 1.5~\cite{reid2024gemini}, GPT-4o, and GPT-4o mini~\cite{openai2024gpt4o}--as well as open-source models such as Phi-3.5-mini-instruct~\cite{abdin2024phi}, Qwen2.5-7B-Instruct, Qwen2.5-72B-Instruct~\cite{yang2024qwen2}, GLM4-9B-Chat~\cite{glm2024chatglm}, Llama-3.1-8B-Instruct, and Llama-3.1-70B-Instruct~\cite{dubey2024llama}.
Gemini Pro 1.5 supports a length of 2M, while the other models support up to 128K.

\paragraph{Training-Free Methods} 
To investigate promising methods, we tested three training-free frameworks specifically designed to efficiently manage long input contexts. 
These frameworks can be applied to any LLM without modification; for our experiments, we used Llama-3.1-8B-Instruct as the backbone LLM.
(1)~StreamingLLM~\cite{xiao2024efficient} leverages the key-value caches of initial tokens within the input context and a finite attention window, capitalizing on the phenomenon where the model's attention heavily ``sinks'' into these initial tokens.
(2)~InfLLM~\cite{xiao2024infllm} selects relevant token sequences from a part of the input that is distant from the current tokens, and it combines these with the initial tokens and local context.
This method has proven effective in existing long-context benchmarks, including InfiniteBench~\cite{zhang-etal-2024-bench} and LongBench~\cite{bai-etal-2024-longbench}.
(3)~MemoRAG~\cite{qian2024memorag} is a retrieval-augmented generation (RAG) framework~\cite{lewis2020retrieval} that consists of a lightweight memory model and a more resource-intensive answer generator. The memory model retrieves answers from a long-context database, which the answer generator then uses as clues to produce the final response.

\subsection{Results}
\label{subsec:results}
Table~\ref{table:main_results} shows the performance of models evaluated on~\ours.
Both open-source and commercial models demonstrated weak overall performance. 
Noticeably, Gemini Pro 1.5 consistently outperformed GPT-4o in our benchmark.
This may be attributed to GPT-4o's 128K context limit, which is significantly lower than the 2M context capacity of Gemini Pro 1.5. 
This result suggests that models with superior long-context handling perform better in our benchmark.

Among the four tasks, the performance gap between the models was most pronounced in the recalling task. 
Gemini Pro 1.5 achieved 69.1\%, while GPT-4o, the second-best model, scored 49.5\%. 
The lowest-performing model managed just 11.7\%. 
Despite the recalling task involving straightforward retrieval queries, the models struggled as the volume of information increased. 
In the summarizing task, the models achieved only moderate performance, which contrasts with the strong performance of recent models in traditional summarization tasks.
Additionally, all models showed room for improvement in both organizing and attributing tasks.

When we applied long-context encoding methods to Llama-3.1-8B-Instruct, there was no performance improvement; in fact, overall performance generally decreased. 
Although these methods have been presented as effective for long-context tasks, they demonstrated limitations in our high-IC tasks, underscoring the need for further research.


\section{Analysis}
\label{section:analysis}

We conducted further analysis to better understand the models' limitations within our benchmark.
We used GPT-4o mini as the main model throughout Sections~\ref{subsec:localglobal} to~\ref{subsec:heatmap}, and all models for Section~\ref{subsec:degeneration}.

\subsection{Comparison of Model Performance Between Low-IC and High-IC Tasks}
\label{subsec:localglobal}

\begin{figure}[t!]
\centering
\includegraphics[width=\columnwidth]{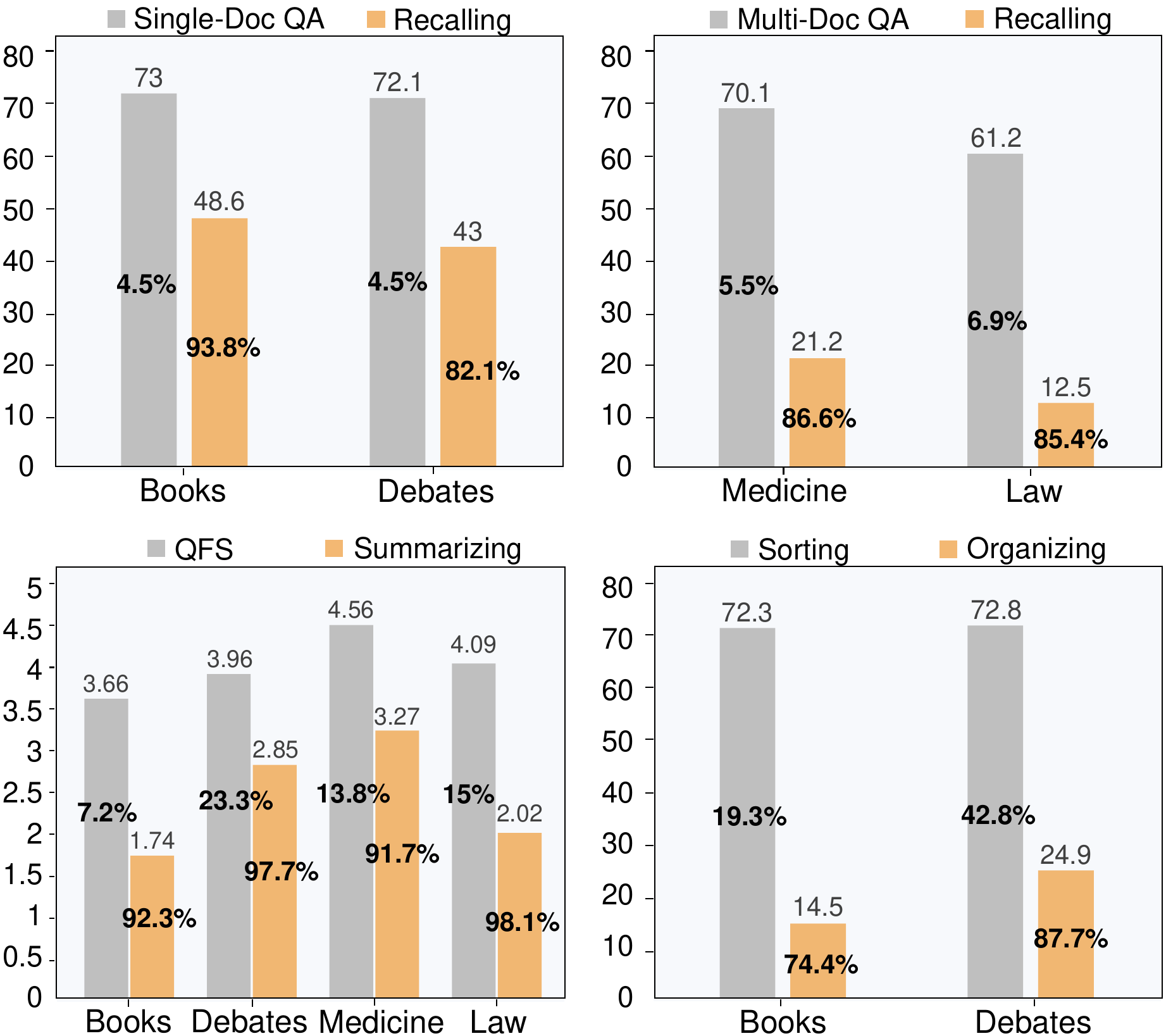}
\caption{
The model's performance on low-IC and high-IC tasks. 
Low-IC tasks were created by generating new queries and answers using the same input context from our benchmark, which are represented by the gray bars on the left side of the graph (please refer to Section~\ref{subsec:localglobal} for details). 
The yellow bars on the right represent high-IC tasks from our benchmark. 
The numbers (\%) displayed in the bar graphs represent the IC values of the tasks.
The y-axis indicates the model performance.
}
\label{fig:localglobal}
\end{figure}

As illustrated in Figure~\ref{fig:motivation}, we observed that model performance can vary significantly based on the amount of information required to answer the query, even when using the same input contexts. 
To explore this further, we constructed a set of low-IC tasks, including traditional NLP tasks such as single-document QA, multi-document QA, and query-focused summarization (QFS). 
Specifically, we extracted one or more chunks from the contexts in our benchmark and asked GPT-4o to generate queries and their corresponding answers.
This method followed a framework similar to that used for developing our benchmark, ensuring high annotation accuracy.
Figure~\ref{fig:localglobal} shows the performance was consistently better on low-IC tasks compared to high-IC tasks. 
This highlights that, even when tasks involve the same documents and similar cognitive demands, information coverage is a key factor that significantly impacts model performance.

\begin{figure}[t!]
\centering
\includegraphics[width=0.75\columnwidth]{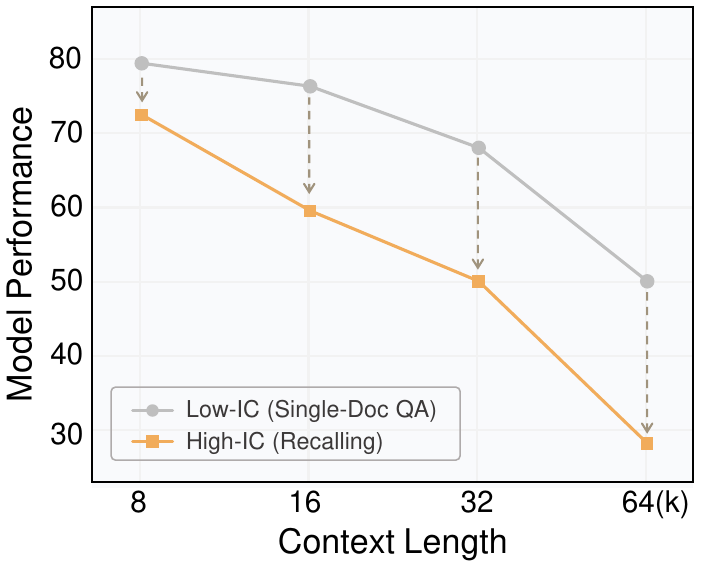}
\caption{
The performance with varied context lengths on low- and high-IC tasks.
We used the single-document QA and recalling tasks from the books and debates domains for low- and high-IC tasks, respectively.
}
\label{fig:context_lengths}
\end{figure}

\subsection{Effect of Context Length}
\label{subsec:difflen}

We examined how model performance in our benchmark is affected by increasing context lengths. 
For this analysis, we utilized recalling queries from the book and debate domain corpora. 
Figure~\ref{fig:context_lengths} illustrates a consistent decline in model performance as the context length increased. 
This trend was also evident in low-IC tasks, as shown in the figure; however, the drop in performance was more pronounced in high-IC tasks compared to low-IC tasks, highlighting their distinct characteristics.

\subsection{Effect of Position of Information}
\label{subsec:heatmap}

\begin{figure}[t!]
\centering
\includegraphics[width=\columnwidth]{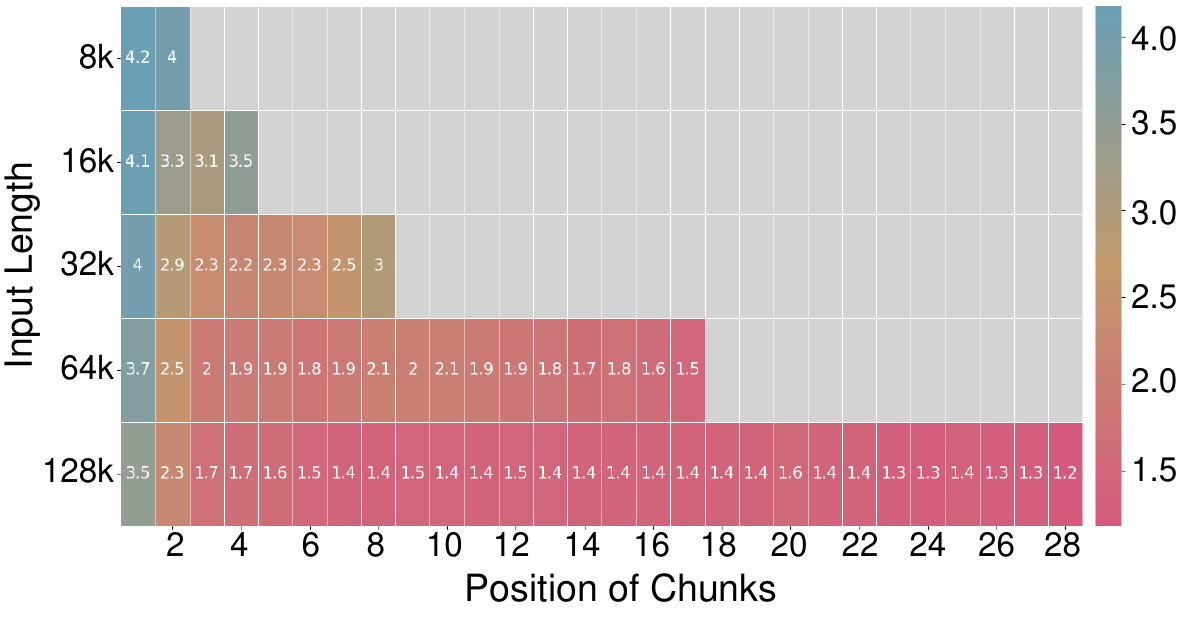}
\caption{
The effect of the position of information in the summarizing task.
The x-axis represents the position of the chunks within the input context, while the y-axis represents the total length of the input context. Blue (red) chunks indicate summaries with high (low) scores.
}
\label{fig:heatmap}
\end{figure}

A recent study demonstrated that when relevant information is located in the middle of the input sequence, both QA and retrieval performance significantly declined~\cite{liu2024lost}. 
To investigate whether a similar position-dependent phenomenon occurs in our tasks, we visualized the model scores for each chunk in the summarizing task.
Figure~\ref{fig:heatmap} shows that the model effectively processed information at the beginning of the provided context, but performance decreased toward the end. 
For context lengths ranging from 8K to 32K, the performance in the middle sections was the lowest, which is consistent with findings from~\citet{liu2024lost}.
When the context length exceeds 32K, we consistently observed a decline in performance as the context length increased.
This might occur because, during decoding, the summaries of the later chunks are influenced by the outputs from the earlier sections, which increases the likelihood of errors in those later parts. 
However, this does not imply that the errors in the later sections are solely due to decoding issues. 
The performance of the earlier sections also decreased as the context length increased, suggesting that the model had inherent limitations in its ability to handle long-context encoding.

\subsection{Error Analysis: Degeneration}
\label{subsec:degeneration}
Through a close examination of model responses, we noticed that models mainly suffer from degeneration~\cite{Welleck2020Neural, DBLP:conf/iclr/HoltzmanBDFC20, Fu_Lam_So_Shi_2021, li2023repetition}, i.e. generation of unreasonably repetitive texts.
Following previous works, we used Rep-\textit{n} and Rep-\textit{r} to quantify this phenomenon across different models (see Appendix~\ref{appendix:appendixF} for details of the metrics). Specifically, we applied Rep-\textit{n} on the Recalling task to account for the portion of repeated \textit{n}-grams, and Rep-\textit{r} on the Summarizing task to account for the portion of repetitive summary snippets. 
As shown in Table~\ref{table:degeneration}, all models exhibited severe degeneration issues across both tasks. This indicates that tasks requiring high information coverage adversely impacted model behavior during generation, ultimately hindering performance.  
Furthermore, notably high repetition scores observed from models such as GPT-4o indicates that degeneration remains a critical issue even for bigger models. Exploring different strategies to mitigate degeneration would be crucial for improving model performance on tasks that demand high information coverage.


\begin{table}[t]
\vspace{2mm}
\centering
\footnotesize
\begin{tabular}{lcc}
\toprule
\multirow{3}{*}{\textbf{Model}} & \multicolumn{1}{c}{\textbf{Recalling}} & \multicolumn{1}{c}{\textbf{Summarizing}}\\
\cmidrule(lr){2-2} \cmidrule(lr){3-3}
 & \multicolumn{1}{c}{\textbf{Rep-n (\%)}} & \multicolumn{1}{c}{\textbf{Rep-r (\%)}} \\
\midrule
\multicolumn{3}{l}{\textit{Proprietary}} \\
\midrule
Gemini Pro 1.5 & \textbf{27.9} & \textbf{43.0}\\
GPT-4o & 90.9 & 65.2\\
GPT-4o mini& 69.8 & 53.8 \\
\midrule
\multicolumn{3}{l}{\textit{Open-Source}} \\
\midrule
Qwen2.5-72B-Instruct& 87.3 & 61.0 \\
Llama-3.1-70B-Instruct& 70.7 & 67.7 \\
\midrule
Qwen2.5-7B-Instruct& 82.9 & 65.5 \\
Llama-3.1-8B-Instruct& 76.8 & 66.0 \\
GLM4-9B-Chat& 89.9 & 63.6 \\
Phi-3.5-mini-instruct & \textbf{55.4} & \textbf{58.3} \\
\bottomrule
\end{tabular}
\caption{
Rep-\textit{n} and Rep-\textit{r} scores across different models on the Recalling and Summarizing tasks.
Please refer to Appendix~\ref{appendix:appendixF} for the details of the metrics.
}
\label{table:degeneration}
\end{table}

\section{Conclusion}
\label{section:conclusion}

In this study, we introduced~\ours, a novel benchmark designed to evaluate LLMs' ability to fully process information in long-context settings. 
We also introduced a new metric, information coverage (IC), to quantify how much of the provided context is required to answer a query.
Compared to existing benchmarks,~\ours~has significantly higher IC values, which underscores its unique aspect. 
Our experiments revealed that current commercial/open-source LLMs and long-context encoding methods perform poorly on our benchmark, emphasizing the need for future research to address long-context processing with high IC.

\section*{Limitations}

We calculated the IC value for long contexts by breaking them into smaller chunks and assessing whether each chunk is necessary to answer the given query. 
However, in cases requiring multi-hop reasoning, some chunks may mislead the evaluator into believing they do not directly contribute to answering the query, even if they are crucial for intermediate reasoning when connected to other chunks. 
Consequently, the evaluator might overlook the potential usefulness of these chunks when evaluating them individually. 
While our study did not thoroughly analyze this aspect, we encourage future research to conduct additional analyses and improvements in this area.

In Figure~\ref{fig:context_lengths}, we show that as the context length increases, the performance of the models on our benchmark significantly declines. 
This drop is likely caused by a combination of difficulties in encoding long contexts and decoding lengthy output sequences. 
We have not analyzed these two factors separately. 
In future work, we aim to develop a more advanced evaluation framework that either fixes the output length while increasing the input context or maintains a constant input context while increasing the output length.

Data contamination and leakage~\cite{magar2022data,xu2024benchmarking} can lead to overestimating model performance and undermine the reliability of benchmarks. 
Recent LLMs are trained on extensive datasets during their pre-training phase, but they do not specify which data was used.
Although we cannot guarantee that we have completely eliminated the possibility of data leakage in our benchmark, we would like to emphasize the steps we have taken to minimize it:
(1) When selecting the book corpus, we chose data from sources where licensing issues were resolved recently (in 2024). This reduces the risk of exposure compared to books that were licensed long ago.
(2) We only collected debate transcripts and legal cases that took place in late 2023 or early 2024.
(3) In the medical domain, we used only the test split of the MS\^{}2 dataset exclusively and did not incorporate any data from the training split.
\section*{Acknowledgements}
This research was supported by (1) the National Research Foundation of Korea (NRF2023R1A2C3004176, RS-2023-00262002), (2) the Ministry of Health \& Welfare, Republic of Korea (HR20C0021), (3) ICT Creative Consilience Program through the Institute of Information \& Communications Technology Planning \& Evaluation (IITP) grant funded by the Korea government (MSIT) (IITP-2025-20200-01819), and (4) Culture, Sports and Tourism R\&D Program through the Korea Creative Content Agency(KOCCA) grant funded by the Ministry of Culture, Sports and Tourism(MCST) in 2023(Project Name: Development of storytelling AI technology for cultural heritage tailored to the various interests of users, Project Number: RS-2023-00220195, Contribution Rate: 100\%).



\bibliography{custom}

\appendix
\clearpage
\renewcommand{\thetable}{\Alph{table}}
\setcounter{table}{0}

\section{Details on Information Coverage}
\label{sec:appendixA}

Below is the instruction used to measure IC.

\vspace{3mm} \scalebox{0.90}{
\hspace{-5mm}
\begin{tcolorbox}[colback=gray!3,colframe=black]
You are given a passage and a query. The given query originally requires a reader to answer based on a longer context. This task splits the long context into multiple small passages, and aims to identify whether a certain passage needs to be taken into account in order to answer the query.\\

\#\#\# Passage:\\
(Note: This passage is a small portion of the original context, and may not be provided in the format specified in the query.)\\
\textcolor{blue}{<passages>}\\

\#\#\# Query:\\
(Note: This query is for evaluation only. DO NOT answer it yourself.)\\
\textcolor{blue}{<query>}\\

\#\#\# Score Rubrics:\\
0 : The passage is an unnecessary part of the original context, which does not need to be taken into account to answer the query.\\
1 : The passage is a necessary part of the original context, which should be taken into account to answer the query.\\

Follow the following format:\\

\# Query Understanding :    \{\{demonstrate a clear understanding of the query and its requirements\}\}\\
\# Passage Understanding : \{\{demonstrate a clear understanding of the passage, focusing on how it relates to the query\}\}\\
\# Assessment : \{\{demonstrate your assessment based on the Score Rubrics provided above\}\}\\
\# Final Score : \{\{a single score only\}\}
\vspace{2mm}
\end{tcolorbox}}

\newpage

\section{Details on Annotation Process}
Below are the instructions used to evaluate models across different tasks. Note that for the Recalling task, the same instructions were used in the annotation process, but with smaller chunks of text.

\label{sec:appendixB}

\subsection{Recalling}
\subsubsection{Books}
\vspace{3mm} \scalebox{0.90}{
\begin{tcolorbox}[colback=gray!3,colframe=black]
\#\#\# Context:\\
\textcolor{blue}{<passages>}\\\\
Now, respond to the instruction.\\
\#\#\# Instruction:\\
List every unique character name that appears in the context. Each name should be a single word. Remove any titles (e.g. Mr., Mrs., Captain, etc.) attached. If a character is mentioned with a full name, list each part of the name separately. Use commas to separate each name. If no such names are present, return "None".\\

Answer:
\end{tcolorbox}}

\subsubsection{Debates}
\vspace{3mm} \scalebox{0.90}{
\begin{tcolorbox}[colback=gray!3,colframe=black]
\#\#\# Context:\\
\textcolor{blue}{<passages>}\\

Now, respond to the instruction.\\
\#\#\# Instruction:\\
List every real name that refers to a person. Remove any titles (e.g. Mr., Mrs., Speaker, Lady, etc.) attached to a name. If a person is mentioned with a full name, list each part of the name separately. Use commas to separate each name. If no such names are present, return "None".\\

Answer:
\end{tcolorbox}}
\newpage
\subsubsection{Medicine}
\vspace{3mm} \scalebox{0.90}{
\begin{tcolorbox}[colback=gray!3,colframe=black]
\#\#\# Context:\\
\textcolor{blue}{<passages>}\\

Now, respond to the instruction.\\
\#\#\# Instruction:\\
List every integer that refers to the count of people (or a group of people) mentioned in the passage. Do not include percentages, proportions, or integers related to non-human entities. If an integer is written in words, convert it to digits. Only return integers, without explanations. Use semicolons to separate each integer. Return "None" if no such integer is mentioned.\\

Answer:
\end{tcolorbox}}
\subsubsection{Law}
\vspace{3mm} \scalebox{0.90}{
\begin{tcolorbox}[colback=gray!3,colframe=black]
\#\#\# Context:\\
\textcolor{blue}{<passages>}\\

Now, respond to the instruction.\\
\#\#\# Instruction:\\
List every number that appears directly in the names of legal citations, including case law numbers, volumes, series, statutes, sections, or codes. Do not include numbers that appear separately from the full citation name, such as standalone paragraph numbers. Remove any special characters that appear with the numbers, and return each number separately. Only return the numbers, separated by semicolons, without explanations or details. If no such numbers are present, return "None".\\

Answer:
\end{tcolorbox}}
\newpage
\subsection{Summarizing}
\vspace{1.5mm} \scalebox{0.90}{
\begin{tcolorbox}[colback=gray!3,colframe=black]
\#\#\# Context:\\
\textcolor{blue}{<passages>}\\

Now, respond to the instruction.\\
\#\#\# Instruction:\\
The context is split into \textcolor{blue}{<num\_sections>} section(s). For each section, write a concise, one-page summary. Prepend the appropriate section header (e.g. <Section 1>) to the summary, and use newlines if there are multiple sections to summarize. Only return the summary, without additional explanations, context, or commentary.
\vspace{1.35cm}
\end{tcolorbox}}
\subsection{Organizing}
\vspace{2mm} \scalebox{0.90}{
\begin{tcolorbox}[colback=gray!3,colframe=black]
\#\#\# Context:\\
\textcolor{blue}{<passages>}\\

Now, respond to the instruction.\\
\#\#\# Instruction:\\
The following options summarize different parts of the given context. Arrange the options in the same order as the events that appear in the context. Only enumerate each option number surrounded by square brackets, without explanations. Use commas to separate your answer.\\

\#\#\# Options:\\
\textcolor{blue}{<options>}\\

Answer:
\vspace{0.9cm}
\end{tcolorbox}}
\newpage
\subsection{Attributing}

\subsubsection{Medicine}
\vspace{3mm} \scalebox{0.90}{
\begin{tcolorbox}[colback=gray!3,colframe=black]
\#\#\# Context:\\
\textcolor{blue}{<passages>}\\

Now, respond to the instruction.\\
\#\#\# Instruction:\\
For each study, assess how the information presented aligns with the selection criteria for systematic reviews focused on clinical outcomes, treatments, or healthcare interventions. Consider whether the study's methodology, findings, or broader implications contribute to the robustness of a systematic review. Label its ID accordingly under one of the following categories:\\
- "Core IDs": Studies that meet the necessary criteria for inclusion in systematic reviews due to their direct contribution to clinical evidence or intervention outcomes.\\
- "Supplementary IDs": Studies that provide additional insights, but may not meet the primary inclusion criteria for systematic reviews.\\

\#\#\# Background:\\
\textcolor{blue}{<background>}\\

Now, label each study under the correct category. Ensure that EVERY study is labeled under at least one category. Use square brackets to surround each ID, without explanations, and separate them by commas. Follow the following format.\\

- Core IDs:\\
- Supplementary IDs:
\vspace{5mm}
\end{tcolorbox}}
\newpage
\subsubsection{Law}
\vspace{7mm} \scalebox{0.90}{
\begin{tcolorbox}[colback=gray!3,colframe=black]
\#\#\# Context:\\
\textcolor{blue}{<passages>}\\

Now, respond to the instruction.\\
\#\#\# Instruction:\\
For the given context and target case, assume that the [citation]s within the context are replaced by the target case, and categorize each SEGMENT based on how the citations could assist in understanding the segment:\\
- "Related Segments": Segments where the "Target Case" provides a clear and direct connection to the citation marks within them, based on legal reasoning, evidence, or laws, making it a valuable reference.\\
- "Supporting Segments": Segments where the "Target Case" may provide some indirect relevance to understanding the context, but it does not have a direct connection to the citation marks in terms of legal reasoning, evidence, or laws.\\

\#\#\# Target Case:\\
\textcolor{blue}{<target\_case>}\\

Now, label each SEGMENT under the correct category. Ensure that EVERY study is labeled under at least one category. Use square brackets to surround each SEGMENT, without explanations, and separate them by commas. Follow the following format.\\

- Related Segments:\\
- Supporting Segments:
\end{tcolorbox}}

\section{Performance per Domain}
\begin{table*}[t]
\centering
\footnotesize
\resizebox{\textwidth}{!}{
\begin{tabular}{l ccc ccc ccc ccc}
\toprule
\multirow{3}{*}{\textbf{Model}} & \multicolumn{3}{c}{\textbf{Books}} & \multicolumn{3}{c}{\textbf{Debates}} & \multicolumn{3}{c}{\textbf{Medicine}} & \multicolumn{3}{c}{\textbf{Law}} \\ 
\cmidrule(lr){2-4} \cmidrule(lr){5-7} \cmidrule(lr){8-10} \cmidrule(lr){11-13}
 & {\textbf{Rec (\%)}} &   {\textbf{Sum.}} & {\textbf{Org (\%)}}   &   {\textbf{Rec.}} & {\textbf{Sum.}}   &   {\textbf{Org}} & {\textbf{Rec.}} &   {\textbf{Sum.}} & {\textbf{Att.}} &    {\textbf{Rec.}} & {\textbf{Sum.}}   &   {\textbf{Att.}}\\
\midrule
\multicolumn{10}{l}{\textit{Proprietary}} \\ \midrule
\multicolumn{1}{l}{Gemini Pro 1.5~\cite{reid2024gemini}}     & {\textbf{61.8}}  &  {\textbf{2.3}} &   {\textbf{38.6}} &  {\textbf{75.4}}  & {2.9}  & {\textbf{62.3}} & {\textbf{79.2}}  & {3.3}  & {\textbf{30.8}} & \textbf{39.5} & 2.5 & 38.1 \\
\multicolumn{1}{l}{GPT-4o~\cite{openai2024gpt4o}}    & {54.4}  & {2.1} & {30.2} & {69.6} & {\textbf{3.2}} & {42.8} &  {44.4} & {\textbf{3.6}} & {30.4} & 11.7 & \textbf{2.6} & \textbf{55.7} \\
\multicolumn{1}{l}{GPT-4o mini~\cite{openai2024gpt4o}}     & {48.6}  &  {1.7} &   {14.5} &  {43.0}  & {2.9}  & {24.9} & {23.4}  & {3.3}  & {22.4} & 12.5 & 2.0 & 42.0\\
\midrule
\textit{Open-Source} \\
\midrule
\multicolumn{1}{l}{Qwen2.5-72B-Instruct~\cite{yang2024qwen2}}      &   {\textbf{53.2}}     &   {\textbf{1.8}}   &   {\textbf{19.9}}  & {\textbf{52.2}}  & {\textbf{3.1}} & {\textbf{31.4}} & {\textbf{49.3}}  & {\textbf{3.5}}  & {\textbf{40.4}} & 14.1 & \textbf{2.6} & 48.0 \\ 
\multicolumn{1}{l}{Llama-3.1-70B-Instruct~\cite{dubey2024llama}}   &  {50.9} & {1.7} & {19.4}  & {38.0}  & {2.5} & {28.5} & {42.0}  & {2.7}  &  {30.5} & \textbf{14.9} & 2.3 & \textbf{55.2} \\ 
\midrule
\multicolumn{1}{l}{Llama-3.1-8B-Instruct~\cite{dubey2024llama}}     &   {29.2}  &   {1.5}   &   {13.6}    &   {11.2}   &   {2.3}   &   {23.1}    & {23.4}      &   {2.6}   &   {24.9} & 10.3 & 2.2 & 36.6 \\ 
\multicolumn{1}{l}{Phi-3.5-mini-instruct~\cite{abdin2024phi}} &  {13.9} &  {1.4}  &  {7.8}  &  {1.0} &  {2.4} &  {19.5}  &  {21.2} & {2.3} & {24.7} & 13.2 & 1.9 & 17.5 \\
\multicolumn{1}{l}{GLM4-9B-Chat~\cite{glm2024chatglm}} &  {33.5} &  {1.6}  &  {16.5}  &  {13.9} &  {2.4} &  {24.6}  &  {18.7} & {2.6} & {35.6} & 12.6 & 2.2 & 9.7 \\
\multicolumn{1}{l}{Qwen2.5-7B-Instruct~\cite{yang2024qwen2}} &  {21.4} &  {1.4}  &  {14.8}  &  {1.5} &  {2.8} &  {20.8}  &  {28.2} & {3.4} & {29.6} & 5.2 & 2.3 & 4.7\\
\midrule
\multicolumn{11}{l}{\textit{Training-Free Methods (built upon Llama3.1-8B-Instruct)}} \\
\midrule
\multicolumn{1}{l}{StreamingLLM~\cite{xiao2024efficient}}   &  {30.3} & {1.2} & {2.9}  & {5.9}  & {1.6} & {1.1} & {22.9}  & {1.8}  &  {11.0} & 7.9 & 1.6 & 9.6 \\ 
\multicolumn{1}{l}{InfLLM~\cite{xiao2024infllm}}      &   {29.1}     &   {1.2}   &   {5.9}  & {9.5}  & {1.8} & {17.2} & {23.6}  & {2.1}  & {12.4} & 6.9 & 1.7 & 13.5 \\ 
\multicolumn{1}{l}{MemoRAG~\cite{qian2024memorag}} &  {29.8} &  {1.2}  &  {15.7}  &  {10.4} &  {1.8} &  {25.0}  &  {25.9} & {1.8} & {17.9} & 7.9 & 1.9 & 14.0 \\

\bottomrule
\end{tabular}
}
\caption{Model performance per domain.}
\label{table:main_stale}
\end{table*}
In Table~\ref{table:main_stale}, we report the performance of models and training-free methods for each domain. 

\section{Implementation Details}
All experiments were done using Nvidia A100 with 80GB memory. For inference, we used vLLM~\cite{kwon2023efficient} for efficiency. We adopted a greedy decoding strategy with temperature set to 0 and top\_p set to 1.0.\\

\section{Validation of Data Quality}
\label{appendix:appendixC}

To ensure the quality of labels generated by GPT-4o, we randomly selected 100 samples from each domain. The resulting annotations were evaluated against GPT-4o-generated labels using the F1 score, as shown in Table~\ref{table:quality_check}.

\begin{table}[H]
\centering
\begin{tabular}{ccc}
\toprule

\textbf{Dataset} & \textbf{F1(\%)}  \\ \midrule

Books & 96.5\\
Debates & 99.0  \\ 
Medicine & 96.8 \\ 
Law & 95.2  \\ 
\bottomrule

\end{tabular}
\caption{Comparison of F1 scores (\%) between GPT-generated labels and human-labeled data across different datasets.}
\label{table:quality_check}
\end{table}

\section{Details on Degeneration Analysis}
We used two metrics to measure degeneration: Rep-\textit{n} and Rep-\textit{r}. Following the notations from~\citet{li2023repetition}, each metric is calculated as follows:\\
\begin{flalign*}
\text{Rep-n} &= 1.0 - \frac{|\text{UniqueNgrams}(x, n)|}{L - n + 1} &&
\end{flalign*}
\vspace{-15pt}
\begin{align*}
\text{Rep-r} = \frac{1}{L} \Big| \Big\{ i \mid 
& (x_i = x_j \wedge x_{i+1} = x_{j+1}, \exists j \neq i) \\
& \hspace{-20pt} \vee (x_i = x_k \wedge x_{i-1} = x_{k-1}, \exists k \neq i) 
\Big\} \Big|
\end{align*}

where $x$, $L$, and \textit{n} refers to the sentence, its length, and the length of \textit{n}-gram within the sentence, respectively. Rep-\textit{n} measures repetition based on the portion of repeated \textit{n}-grams, whereas Rep-\textit{r} quantifies repetition based on the portion of repeated snippets measured by sentence length.

\label{appendix:appendixF}

\end{document}